\documentclass{ecai-template/ecai} 

\usepackage{latexsym}
\usepackage{amssymb}
\usepackage{amsmath}
\usepackage{amsthm}
\usepackage{booktabs}
\usepackage{enumitem}
\usepackage{graphicx}
\usepackage{color}
\usepackage{multirow}
\usepackage{subcaption}
\usepackage{hyperref}
\usepackage{cleveref}
\usepackage{algorithm}
\usepackage[noend]{algorithmic}
\usepackage[switch]{lineno}
\usepackage[separate-uncertainty=true,multi-part-units=single]{siunitx}

\newcommand{\BibTeX}{B\kern-.05em{\sc i\kern-.025em b}\kern-.08em\TeX}

\begin{document}


\begin{frontmatter}


\paperid{1094} 


\title{Harnessing Orthogonality to Train Low-Rank Neural Networks}

\author[a]{\fnms{Daniel}~\snm{Coquelin}\orcid{0000-0001-8552-5153}\thanks{Corresponding Author. Email: daniel.coquelin@kit.edu.}}
\author[a]{\fnms{Katharina}~\snm{Fl\"ugel}}
\author[a]{\fnms{Marie}~\snm{Weiel}}
\author[a]{\fnms{Nicholas}~\snm{Kiefer}}
\author[a]{\fnms{Charlotte}~\snm{Debus}}
\author[a]{\fnms{Achim}~\snm{Streit}}
\author[a]{\fnms{Markus}~\snm{G\"otz}}
 
\address[a]{Scientific Computing Center (SCC), Karlsruhe Institute of Technology (KIT),\\ 76344 Eggenstein-Leopoldshafen, Germany}


\begin{abstract}
This study explores the learning dynamics of neural networks by analyzing the singular value decomposition (SVD) of their weights throughout training. 
Our investigation reveals that an orthogonal basis within each multidimensional weight's SVD representation stabilizes during training.
Building upon this, we introduce Orthogonality-Informed Adaptive Low-Rank (OIALR) training, a novel training method exploiting the intrinsic orthogonality of neural networks.
OIALR seamlessly integrates into existing training workflows with minimal accuracy loss, as demonstrated by benchmarking on various datasets and well-established network architectures. 
With appropriate hyperparameter tuning, OIALR can surpass conventional training setups, including those of state-of-the-art models. 
\end{abstract}

\end{frontmatter}


\section{Introduction}

How does a neural network learn?
Mathematically, its weights are adjusted iteratively, most commonly using the back-propagated gradients of the loss function to minimize the difference between predicted outputs and actual targets.
Yet, we have only a limited understanding of what characteristics are being learned.
In this work, we show that this optimization process imposes a structure on the network and present a way to exploit it.
State-of-the-art neural networks have long existed at the edge of what is computationally possible.
Hence, there has long been a desire to reduce the size of networks by sparsification or compression.
It has been repeatedly shown that both of these methods can be greatly effective given proper training procedures.

As most tensors can be expressed as low-rank approximations, these methods are often used to compress neural networks~\citep{deng2020compressionsurevy}.
A low-rank approximation factorizes a full-rank matrix $\boldsymbol{M}$ into two or more matrices where the inner dimension $r$ is smaller than the original matrix's dimensions. 
Formally, the factorization is defined as $\boldsymbol{M}_{m\times n}=\boldsymbol{A}_{m\times r}\boldsymbol{B}_{r\times n}$ with $r < \min(m,n)$.
Low-rank network representations can have a variety of uses, from reducing the computational complexity of training~\citep{schotthofer2022lowrank} or validation~\citep{singh2019pp}, to fine-tuning large language models in a resource-efficient manner, e.g., LoRA~\citep{hu2022lora}.

Singular value decomposition (SVD) is a popular method for finding a, theoretically exact, low-rank representation of a given matrix.
SVD factorizes a matrix into an orthogonal basis $\boldsymbol{U}$, an orthogonal cobasis $\boldsymbol{V}$, and a diagonal matrix of the singular values sorted in descending order $\boldsymbol{\Sigma}$, as $\boldsymbol{M}_{m\times n}=\boldsymbol{U}_{m\times r}\boldsymbol{\Sigma}_{r\times r} \boldsymbol{V}_{r\times n}^T$.
To produce a low-rank approximation, singular values and their corresponding basis vectors can be removed.
However, as larger singular values are removed, the approximation quality diminishes.

The power of SVD lies in its ability to identify the principal components of a matrix, i.e., the directions along which it varies most.
These principal components are captured by the orthogonal bases $\boldsymbol{U}$ and $\boldsymbol{V}$.
We posit that these components are learned in the initial phases of training, which allows for network compression in later training stages. In this work, we
\begin{itemize}
    \item show evidence that the orthogonal basis $\boldsymbol{U}\boldsymbol{V}^T$ of each of a network's multidimensional weights stabilizes during training;
    \item propose Orthogonality-Informed Adaptive Low-Rank (OIALR) neural network training, a novel training method harnessing this finding;
    \item demonstrate that OIALR seamlessly integrates into existing training workflows with minimal accuracy loss by means of benchmarks on multiple datasets, data modalities, well-known and state-of-the-art network architectures, and training tasks;
    \item show that OIALR can outperform conventional full-rank and other low-rank training methods.
\end{itemize}
For simplicity, we will refer to the orthogonal basis $\boldsymbol{U}\boldsymbol{V}^T$ of each of a network's multidimensional weights as the network's orthogonal bases.
We provide public implementations of the most common layer types, a method to wrap arbitrary model architectures for any learning task, and the OIALR training method \href{https://github.com/Helmholtz-AI-Energy/oialr}{here}. 

\section{Related work}

Several methods exist for obtaining a reduced-size network, including pruning, which involves the removal of unimportant weights in a model; compression, which reduces the size of weights via low-rank approximations; quantization, which reduces the number of bits used to store the weights; and sparse training, where the network is trained to be sparse from initialization~\citep{xu2023compresssurvey}.

Convolution layers famously contain kernel~\citep{hssayni2022_krr-cnn}, channel~\citep{he2017cp}, and filter~\citep{singh2019pp} redundancies which can be pruned.
These methods are referred to as structured pruning methods.
Although convolution layers are the use case of these methods, other methods have been found for different layer types~\citep{wang2020structured}.
In contrast, unstructured pruning methods prune individual weights across all layer types~\citep{lin2019gal}.

Low-rank compression methods often employ SVD. 
To the best of our knowledge, the earliest application of SVD in neural network compression is the SVD-NET~\citep{psichogios1994svd-net}.
This method decomposes the weights of a simple neural network with SVD and trains the $\boldsymbol{U}$, $\boldsymbol{\Sigma}$, and $\boldsymbol{V}$ matrices.
This approach has been successful and has been shown to improve generalization, i.e. reducing overfitting during training~\citep{winata2020lowranktrans-speech,phan2020stablecnn,cahyawijaya2021greenformer}.

Given the abundance of pre-trained models today, there is a strong inclination to apply a low-rank approximation after training.
Although this often leads to significant performance degradation~\citep{xu2019train}, fine-tuning pre-trained models for specific use cases using low-rank approximations has proven to be effective~\citep{mahabadi2021compacter}.
For instance, LoRA~\citep{hu2022lora} demonstrated that large language models (LLMs) can be adapted to specialized use cases by adding a low-rank weight component alongside the originally trained weights.

The most common approach for training low-rank models from scratch is to train all matrices' low-rank approximations~\citep{ceruti2021rank-adaptive,hsu2022svdllm,guo2023conformerlowrank}. 
As starting the training in low rank can be detrimental to network accuracy~\citep{waleffe2020pcanets,bejani2020lowrank-regularize}, many methods transition to a low-rank representation later or slowly reduce the inner-rank during training~\citep{xu2023compresssurvey}.

A matrix's orthogonal basis can be viewed as the coordinate system that the matrix operates in, making it extremely useful for explaining the inner working of modern black-box neural networks.
Intuitively, orthogonal networks are beneficial from an explainability viewpoint.
ExNN~\citep{yang2021exnn} utilized methods to preserve projection orthogonality during training to improve interpretability.
Similarly, the Bort optimizer~\citep{zhang2023bort} aims to improve model explainability using boundedness and orthogonality constraints.

While many low-rank approximations utilize orthogonality, low-rank training methods frequently do not maintain it.
Methods that respect orthogonality, show increased network performance~\citep{povey2018semiorthogonal,schotthofer2022lowrank}.
In DLRT~\citep{schotthofer2022lowrank}, each of the three matrices in a model's singular-value decomposed weights is trained in a separate forward-backward pass while maintaining orthogonality.

\section{Observing orthogonality in neural network training}
\label{sec:basis}

\begin{figure*}[tb]
     \centering
     \begin{subfigure}[t]{0.495\textwidth}
         \centering
        \includegraphics[width=\linewidth]{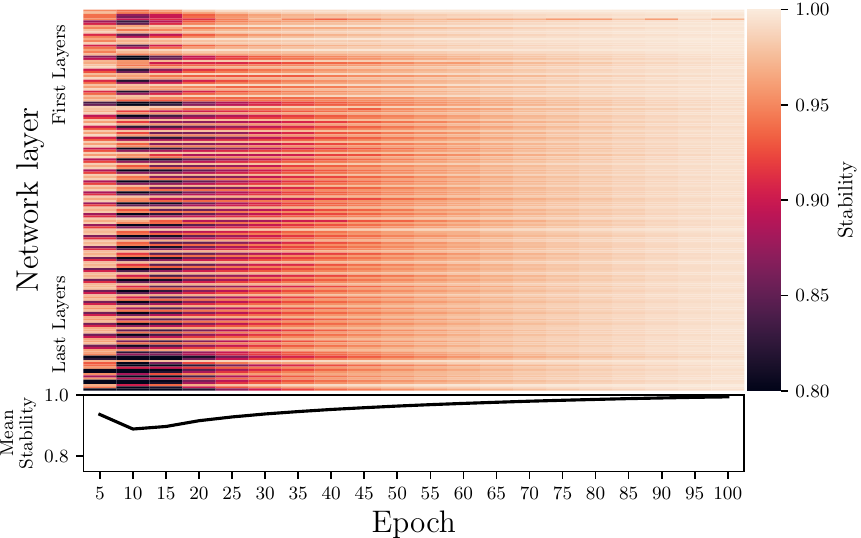}
        \caption{ResNet-RS 101, \textsl{Stability}, see~\Cref{eq:stability}}
        \label{fig:stab-resnet}
     \end{subfigure}
     \hfill
     \begin{subfigure}[t]{0.495\textwidth}
         \centering
          \includegraphics[width=\linewidth]{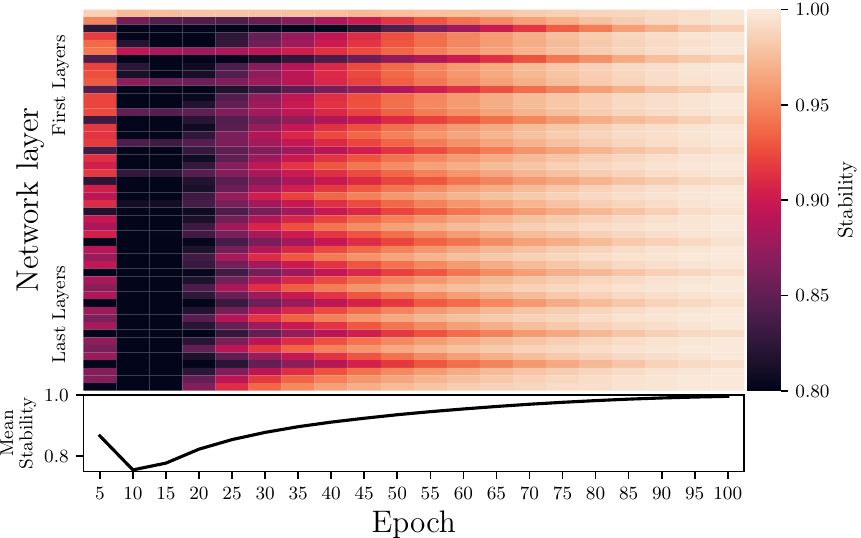}
          \caption{Vision Transformer B/16, \textsl{Stability}, see~\Cref{eq:stability}\vspace{0.4cm}}
          \label{fig:stab-vit}
     \end{subfigure}
     \begin{subfigure}[t]{0.495\textwidth}
         \centering
        \includegraphics[width=\linewidth]{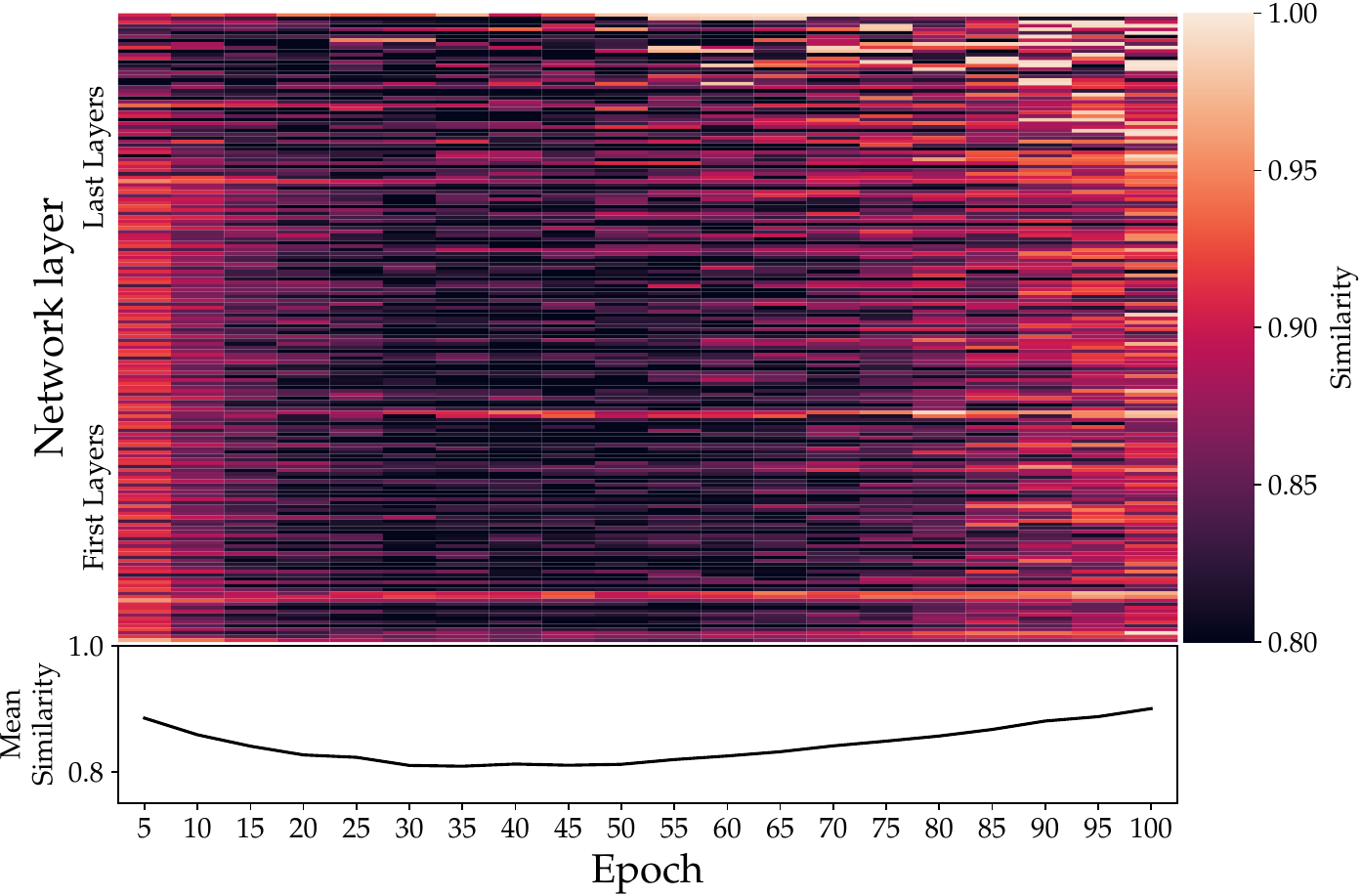}
        \caption{ResNet-RS 101, Euclidean similarity, see~\Cref{eq:dist}}
        \label{fig:dist-resnet}
     \end{subfigure}
     \hfill
     \begin{subfigure}[t]{0.495\textwidth}
         \centering
          \includegraphics[width=\linewidth]{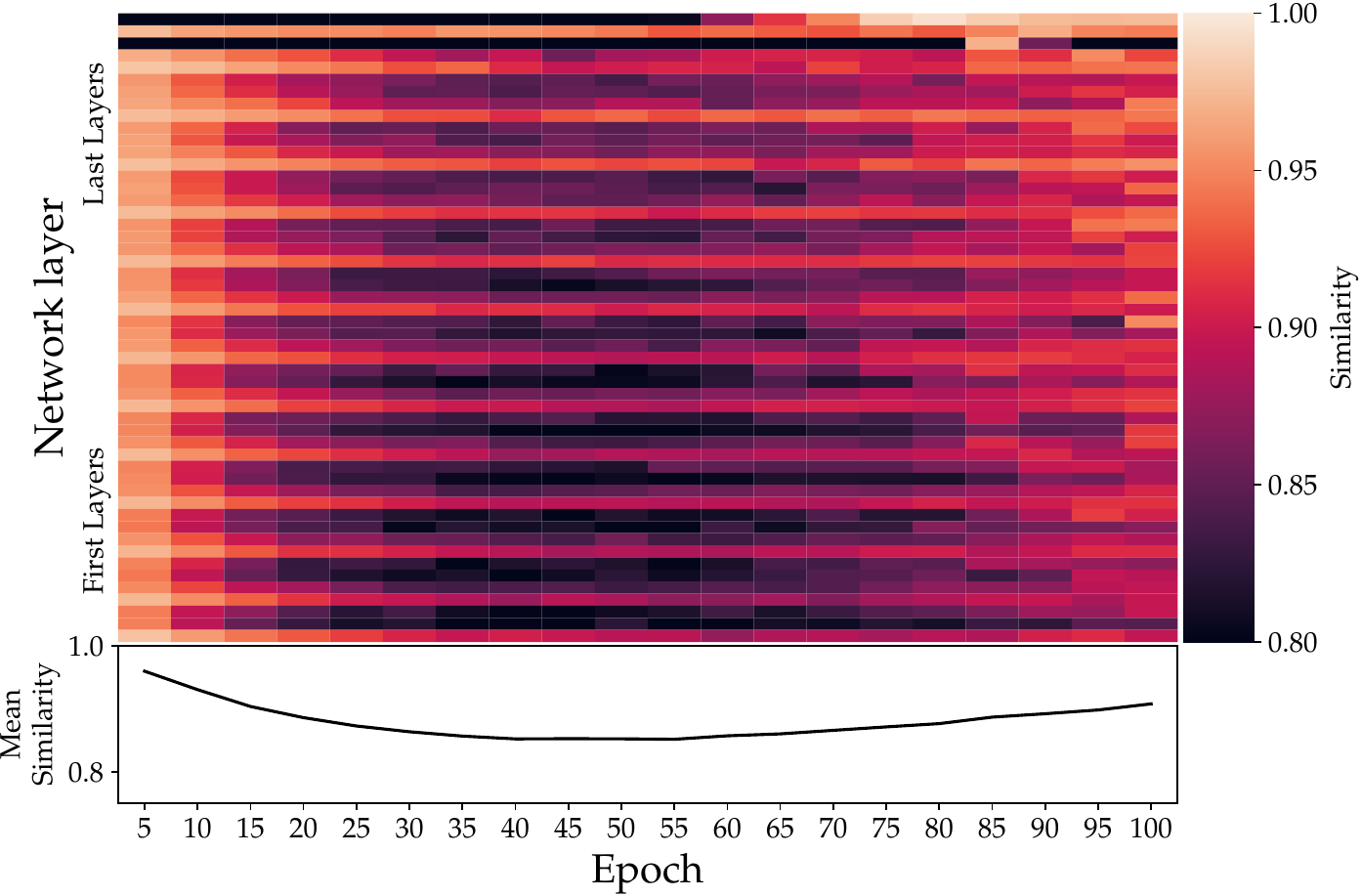}
          \caption{Vision Transformer B/16, Euclidean similarity, see~\Cref{eq:dist}
          \vspace{0.4cm}
          }
          \label{fig:dist-vit}
     \end{subfigure}
    \caption{
    Analysis of the linear mixing Euclidean similarity and orthogonal basis \textsl{Stability} for ResNet and ViT models during ImageNet-2012 training. 
    \textsl{Stability} is defined by \Cref{eq:stability} and Euclidean similarity is defined by \Cref{eq:dist}, higher denotes less changes between steps for both. 
    Both metrics compare the network's current parameters with those of five epochs prior.
    The x-axis denotes the training epoch, and the y-axis denotes the network layer (input layers at the top). 
    Mean stability and similarity are shown below each heatmap.
     \\~}
    \label{fig:stability}
\end{figure*}  

The work of \citet{schotthofer2022lowrank} formulates the training of a low-rank neural network as a continuous-time gradient flow.
They use low-rank numerical integrators for matrix Ordinary Differential Equations (ODEs) to create a training process which uses three forward and backward passes for each batch to train a low-rank neural network.
This formulation of training with ODEs is quite appealing as it can accurately show the impact factorizations have on the gradients of the parameters that they represent.

We begin from the same formulation but significantly simplify the low-rank training process by showing that a network parameter's orthogonal component stabilizes during the training.
Let us consider a single weight matrix of network layer, $k$, of shape ${m\times n}$ at step $t$ of training, $\boldsymbol{W}_k(t)\in \mathcal{M}_{r_k}$ where $\mathcal{M}_{r_k}$ is the manifold of matrices with rank $r_k$.
The other weights are fixed in time and treated as constants for gradient calculations.
From \citet{schotthofer2022lowrank}, the training can then be formulated as a continuous process:
\begin{equation}
    \resizebox{.9\hsize}{!}{$
        \text{min}\{ \| \dot{\boldsymbol{W}_k}\left(t\right) + \nabla_{\boldsymbol{W}_k}\mathcal{L}\left(\boldsymbol{W}_k\left(t\right)\right)\| : \dot{\boldsymbol{W}_k}\left(t\right) \in \mathcal{T}_{\boldsymbol{W}_k\left(t\right)}\mathcal{M}_{r_k} \}
    $}
    \label{eq1}
\end{equation}
where $\mathcal{L}$ is the loss function, $\mathcal{T}_{\boldsymbol{W}_k\left(t\right)}\mathcal{M}_{r_k}$ is the tangent space of $\mathcal{M}_{r_k}$ at position $\boldsymbol{W}_k\left(t\right)$, and $\dot{\boldsymbol{W}_k}\left(t\right)$ denotes the temporal derivative.
This can be expressed as a Galerkin condition:
\begin{equation} \label{eq:galer}
    \resizebox{.9\hsize}{!}{$
        \langle \dot{\boldsymbol{W}_k}(t) + \nabla_{\boldsymbol{W}_k}\mathcal{L}(\boldsymbol{W}_k(t)), \delta \boldsymbol{W}_k \rangle = 0 \quad \forall\delta \boldsymbol{W}_k \in \mathcal{T}_{\boldsymbol{W}_k(t)}\mathcal{M}_{r_k}
    $}
\end{equation}
where $\delta\boldsymbol{W}_k$ is an element of the tangent space $\mathcal{T}_{\boldsymbol{W}_k\left(t\right)}\mathcal{M}_{r_k}$.

Using the SVD decomposition, $\boldsymbol{W}_k = \boldsymbol{U}_k \boldsymbol{\Sigma}_k \boldsymbol{V}_k^\top$, this element can be described as:
\begin{equation}\label{eq:del-w}
    \delta\boldsymbol{W}_k = \delta \boldsymbol{U}_k \boldsymbol{\Sigma}_k \boldsymbol{V}_k^\top + \boldsymbol{U}_k \delta \boldsymbol{\Sigma}_k \boldsymbol{V}_k^\top + \boldsymbol{U}_k \boldsymbol{\Sigma}_k \delta \boldsymbol{V}_k^\top
\end{equation}
where $\delta \boldsymbol{U}_k$ and $\delta \boldsymbol{V}_k$ are elements of the tangent space of the Stiefel manifold with $r_k$ orthonormal columns at the points $\boldsymbol{U}_k$ and $\boldsymbol{V}_k$, and $\boldsymbol{\Sigma}_k$ is a matrix of shape $r_k \times r_k$.

Lets consider the QR decomposition of $\boldsymbol{W}_k$.
A small change to $\boldsymbol{W}_k$ can result in changes to the orthogonal basis $\boldsymbol{Q}_k$, the mixing matrix $\boldsymbol{R}_k$, or both.
However, given that SGD makes small iterative steps, we postulate that large-scale changes to the basis happen predominately at low step counts.
In short, we hypothesise that the orthogonal component of $\boldsymbol{W}_k$ stabilizes during the early stages of training.

Most of a network's weights can be represented by tall-and-skinny matrices, i.e. $m > n$. 
For these weights, the full $\boldsymbol{Q}$ matrix is of size $\mathcal{O}(m^2)$, making it memory inefficient to track $\boldsymbol{Q}$ over multiple training steps.
To test our hypothesis, we use the semi-orthogonal bases $\boldsymbol{U}_k\boldsymbol{V}_k^\top$ as determined by the compact SVD of $\boldsymbol{W}_k$ as the orthogonal component.
This semi-orthogonal matrix is more memory efficient, $\mathcal{O}\left(mn\right)$, and uniquely determined by $\boldsymbol{W}_k\left(\boldsymbol{W}_k^T \boldsymbol{W}_k\right)^{-1/2}$.
To track how the orthogonal component of $\boldsymbol{W}_k$ changes throughout training we define the \textsl{Stability} between timesteps $i$ and $j$ as
\begin{equation}
    S_{k,ij} = \frac{\mathrm{tr}\left( \left(\boldsymbol{U}_k\boldsymbol{V}_k^\top\right)_{i} \left(\boldsymbol{V}_k\boldsymbol{U}_k^\top\right)_{j} \right)}{m}
    \label{eq:stability}
\end{equation}
where $\mathrm{tr}$ is the trace and the product $\left(\boldsymbol{U}_k\boldsymbol{V}^\top_{k}\right)_i$ is orthogonal component of $\boldsymbol{W}_k$ at timestep $i$.
This metric ranges from zero to one, where a \textsl{Stability} of zero indicates two completely different bases and a \textsl{Stability} of one indicates two identical bases.
To track the mixing matrix, $\boldsymbol{R}_k$, we use a form of Euclidean similarity 
\begin{equation} \label{eq:dist}
    D_{k,ij} = 1- \sqrt{\frac{\left(\boldsymbol{R}_{k,i} - \boldsymbol{R}_{k,j} \right)^2 }{mn}}
\end{equation}
where $\boldsymbol{R}_{k,i}$ and $\boldsymbol{R}_{k,i}$ are the $\boldsymbol{R}$ matrices for $\boldsymbol{W}_k$ at two different timsteps $i$ and $j$.
This similarity metric is chosen to maintain the same scale and behavior as \textsl{Stability}. 

\Cref{fig:stability} shows how both the mixing matrix and the orthogonal component of the weights evolve during the training of two vastly different network architectures, ResNet-RS 101~\citep{bello2021resnetrs} and the VisionTransformer (ViT) B/16~\citep{dosovitskiy2021vit}, on ImageNet-2012~\citep{russakovsky2015imagenet}.
In both cases (\Cref{fig:stab-resnet,fig:stab-vit}) the \textsl{Stability} decreases in the first ten epochs, i.e. the parameters' basis vectors are changing, as the networks move away from their random initialization.
Then, the \textsl{Stability} converges towards its maximum value of one, indicating that the current basis and the previous iteration are becoming more aligned.
This is in sharp contrast to the similarity plots (\Cref{fig:dist-resnet,fig:dist-vit}), where the largest changes to the mixing occur towards the middle of training, while at the beginning and end of training the changes to the mixing are smaller.
This finding aligns itself nicely with fact that gradients converge in direction during training~\cite{ji2020directional}.

Using the information from \Cref{fig:stability}, we can assume that in the later stages of training $\delta \boldsymbol{U}_k$ and $\delta \boldsymbol{V}_k$ tend to zero; and \Cref{eq:del-w} reduces to:
\begin{equation}\label{eq:new-del-w}
    \delta\boldsymbol{W}_k = \boldsymbol{U}_k \delta \boldsymbol{\Sigma}_k \boldsymbol{V}_k^\top
\end{equation}
and the time derivative of $\boldsymbol{W}_k$ becomes
\begin{equation}
    \dot{\boldsymbol{W}}_k = \frac{d}{dt}\left\{\boldsymbol{U}_k\boldsymbol{\Sigma}\boldsymbol{V}_k^\top \right\} = \boldsymbol{U}_k \dot{\boldsymbol{\Sigma}}_k \boldsymbol{V}_k^\top.
\end{equation}
Thus, later in training, the Galerkin condition, \Cref{eq:galer}, is now
\begin{equation} \label{eq:galer2}
    \resizebox{.9\hsize}{!}{$
        \langle \boldsymbol{U}_k \dot{\boldsymbol{\Sigma}}_k \boldsymbol{V}_k^\top + \nabla_{\boldsymbol{W}_k}\mathcal{L}(\boldsymbol{W}_k(t)), \delta \boldsymbol{W}_k \rangle = 0 \quad \forall\delta \boldsymbol{W}_k \in \mathcal{T}_{\boldsymbol{W}_k(t)}\mathcal{M}_{r_k}
    $}
\end{equation}
and as we can treat $\boldsymbol{U}_k$ and $\boldsymbol{V}_k$ as constants, 
\begin{equation} \label{eq:new-sdot}
    \langle \dot{\boldsymbol{\Sigma}}_k + \boldsymbol{U}_k^\top \nabla_{\boldsymbol{W}_k}\mathcal{L}(\boldsymbol{W}_k(t))\boldsymbol{V}_k, \delta \boldsymbol{\Sigma}_k \rangle = 0
\end{equation}
finally, we arrive at
\begin{equation}
    \dot{\boldsymbol{\Sigma}}_k = -\boldsymbol{U}_k^\top \nabla_{\boldsymbol{W}_k} \mathcal{L}(\boldsymbol{W}_k(t))\boldsymbol{V}_k.
\end{equation}
This allows for a massive simplification of low-rank training: only training on the low-rank matrix $\boldsymbol{\Sigma}_k$ while using low-rank numerical integrators for matrix ODEs as shown by \citet{schotthofer2022lowrank}.

\section{Orthogonality-Informed Adaptive Low-Rank Training}
\label{sec:oialr}
To harness the stabilization of the orthogonal bases in neural network training, we present a novel algorithm that reduces the number of trainable parameters while maintaining both accuracy and overall time-to-train, unlike most previous methods which focused on either one or the other~\citep{xu2023compresssurvey}.

As shown in \Cref{fig:stability}, most layers’ bases do not stabilize before a few epochs have passed. 
Therefore, we start training in a traditional full-rank scheme. 
After a number of iterations $d$, a hyperparameter of the algorithm, we transition the network's multidimensional weights to their $\boldsymbol{U}\Sigma\boldsymbol{V}^T$ representation via their SVD. 
Experimentally, we found that the delay should be one third of the total number of iterations.
At this point, we no longer train $\boldsymbol{U}$ and $\boldsymbol{V}^T$ with backpropagation but \emph{train only the square matrix $\boldsymbol{\Sigma}$}.
After a specified number of training steps $\nu$, the bases $\boldsymbol{U}$ and $\boldsymbol{V}^T$ are updated by extracting the new bases from the trained $\boldsymbol{\Sigma}$ matrix using an SVD of $\boldsymbol{\Sigma}$, as outlined in Algorithm 1.
\begin{algorithm}[tb]
    \caption{Updating a weight matrix's basis and cobasis, $\boldsymbol{U}$ and $\boldsymbol{V}$, from its actively trained singular value matrix $\boldsymbol{\Sigma}$.
    $\boldsymbol{U}$, $\boldsymbol{\Sigma}$, and $\boldsymbol{V}$ comprise the current $\boldsymbol{U}\boldsymbol{\Sigma}\boldsymbol{V}^T$ representation of a weight matrix.} 
    \label{algo:update}
	\textbf{Inputs}: Frozen bases $\boldsymbol{U}$ and $\boldsymbol{V}$, trained $\boldsymbol{\Sigma}$ matrix\;
 
    \begin{algorithmic}[1] 
        \STATE $\boldsymbol{U}^{'}, \boldsymbol{\Sigma}^{'}, \boldsymbol{V}^{'T} \gets \texttt{SVD}(\boldsymbol{\Sigma})$
        \STATE $\boldsymbol{U} \gets \boldsymbol{U}\boldsymbol{U}^{'}, \boldsymbol{V}^{T} \gets \boldsymbol{V}^{'T}\boldsymbol{V}^{T}, \boldsymbol{\Sigma} \gets \boldsymbol{\Sigma}^{'}$ 
    \end{algorithmic}
\end{algorithm}
After the basis $\boldsymbol{U}$ and cobasis $\boldsymbol{V}^T$ are updated, a new inner rank is found by removing the singular values whose absolute magnitude is less than $\beta$ times the largest singular value in the current $\boldsymbol{\Sigma}$, where $\beta$ is a hyperparameter that defaults to $0.1$.
As the first layers of the network are generally unstable for longer, the update of $\boldsymbol{U}$ and $\boldsymbol{V}$ is only applied to the network's last $\ell = L\cdot \alpha \cdot u$ layers, where $L$ is the number of network layers, $\alpha$ is a hyperparameter defaulting to $0.1$, and $u$ is the number of already completed updates.
This process repeats until the end of training.
Optionally, the first or last layers can be excluded from low-rank training depending on the use case.
We provide an outline of our Orthogonality-Informed Adaptive Low-Rank (OIALR) training in Algorithm 2.

The first transition to the $\boldsymbol{U}\Sigma\boldsymbol{V}^T$ representation will almost certainly utilize more memory than in its traditional state.
For example, a traditional weight matrix has $\mathcal{O}(mn)$ elements while the $\boldsymbol{U}\Sigma\boldsymbol{V}^T$ representation in OIALR has $\mathcal{O}\left( r (m+r+n)\right)$ elements, where $r$ is the inner rank.
As training progresses, the number of `useful' basis vectors is expected to decrease, resulting in a reduction in the network's size, given that the network is trained appropriately.

\begin{algorithm}[tb]
    \caption{OIALR training method}
    \label{algo:train}
	\textbf{Inputs}: Model $M$, training steps $t_{\max}$, delay steps $d$, low-rank update frequency $\nu$, singular value cutoff fraction $\beta$, percentage of layers in each low-rank update step size $\alpha$ \\
    \textbf{Parameter}: $L \gets \text{Number of possible low-rank weights in } M$\\
    \textbf{Parameter}: $\ell \gets L \cdot \alpha$
    \begin{algorithmic}[1]
        \FOR{$t\gets1$ {\bfseries to} $t_{\max}$}
            \IF{$t < d$}
                \STATE Train full-rank network.
            \ELSIF{$t = d$}
                \STATE Convert network to $\boldsymbol{U}\boldsymbol{\Sigma}\boldsymbol{V}^T$ representation.
            \ELSIF{$t \texttt{ mod } \nu = 0$}
                \FOR{$i\gets L - \ell$ {\bfseries to} $L$}
                    \STATE Update $\boldsymbol{U}\boldsymbol{\Sigma}\boldsymbol{V}^T$ rep. $i$ with Algorithm 1.
                    \STATE Remove singular values $< \beta \cdot \texttt{max}(\boldsymbol{\Sigma}_i)$
                    \STATE Reshape $\boldsymbol{U}_i$, $\boldsymbol{V}_i$, $\boldsymbol{\Sigma}_i$, and optimizer states.
                \ENDFOR
                \STATE $\ell \gets \ell + L \cdot \alpha$
            \ELSE
                \STATE Train network's $\boldsymbol{U}\boldsymbol{\Sigma}\boldsymbol{V}^T$ representations ($\boldsymbol{\Sigma}$).
            \ENDIF
        \ENDFOR
    \end{algorithmic}
\end{algorithm}

\section{Experiments}

To evaluate the effectiveness of our OIALR training approach, we conducted extensive experiments using different neural network architectures and datasets. 
Our primary focus was to demonstrate its effectiveness in terms of reducing the number of trainable parameters while maintaining or enhancing network performance and training time.

In our first experiment, we aim to understand what a typical researcher would experience by applying OIALR directly to a conventional and well-known neural network setup for a computer vision problem (see \Cref{sec:vit}).
Next, we compare OIALR to other popular low-rank and sparse methods (see \Cref{sec:comp}).
To see how OIALR performs on a real-world application, we investigate its performance in time-series forecasting with the Autoformer~\citep{wu2021autoformer}, which was deployed at the 2022 Winter Olympics.

Up to this point, the hyperparameters (HPs) of the training methods are the same for both the baseline and OIALR methods.
Given that OIALR dynamically alters the network structure during training, we expect the optimal HPs for OIALR training to vary from those used for full-rank training.
For the final two experiments in \Cref{sec:cifar,sec:ettm2}, we also determine more well-suited HPs for OIALR using \texttt{Propulate}~\citep{taubert2023massively}, an asynchronous evolutionary optimization package shown to be effective for neural architecture searches~\citep{coquelin2021rsapplication}.
We report `Compression' and `Trainable parameters' as percentages relative to the conventional model.
For example, a compression of 80\% signifies 20\% fewer total parameters than the traditional model.
Non-trainable parameters for OIALR-trained models are dominated by the $\boldsymbol{U}$ and $\boldsymbol{V}$ bases, while traditional models tend to have fewer non-trainable parameters like the running average in a batch normalization layer.

To demonstrate how our method would perform on real-world use cases, our experiments used state-of-the-art techniques and models, including strong image transforms~\citep{touvron2021augments}, dropout~\citep{srivastava2014dropout}, learning rate warm-up~\citep{gotmare2018lrwarmup}, and cosine learning rate decay~\citep{loshchilov2017cosine}, implemented as per \citep{timm_2023}.
All networks were trained using the AdamW~\citep{loshchilov2018adamw} optimizer. 
Complete sets of HPs are included in 
\Cref{sec:hypers}. 
Results represent the average of three runs with distinct random seeds.

\subsection{Computational environment}

We ran all experiments on a distributed-memory, parallel hybrid supercomputer. 
Each compute node is equipped with two 38-core Intel Xeon Platinum 8368 processors at \SI{2.4}{\giga\hertz} base and \SI{3.4}{\giga\hertz} maximum turbo frequency, \SI{512}{\giga\byte} local memory, a local \SI{960}{\giga\byte} NVMe SSD disk, two network adapters, and four NVIDIA A100-40 GPUs with \SI{40}{\giga\byte} memory connected via NVLink. 
Inter-node communication uses a low-latency, non-blocking NVIDIA Mellanox InfiniBand 4X HDR interconnect with \SI{200}{\giga\bit/\second} per port. 
All experiments used Python 3.10.6 with \texttt{CUDA}-enabled \texttt{PyTorch} 2.0.0~\citep{paszke2018pytorch}. 

\subsection{Vision Transformer on ImageNet-2012}
\label{sec:vit}
\begin{table*}[t]
\centering
\caption{Training ViT-B/16 on ImageNet-2012 with and without OIALR. Hyperparameters are identical in both cases.}
\label{tab:imagenet}
\begin{tabular}{@{}c l c c c c c c @{}}
\toprule
 & \begin{tabular}[c]{@{}c@{}}Training\\method \end{tabular} & Loss & Top-1 acc. & Top-5 acc. & Time to train & Compression & \begin{tabular}[c]{@{}c@{}}Trainable \\parameters \end{tabular}    \\ \midrule
\multirow{2}{*}{\rotatebox[origin=c]{0}{ViT-B/16}} & Baseline & \textbf{2.16} & \textbf{71.64 \%} & \textbf{89.18 \%} & 3.29 \si{\hour} & --- & --- \\
 & OIALR  & 2.20 & 70.30 \% & 88.73 \% & \textbf{3.26 \si{\hour}} & \textbf{98.95 \%} & \textbf{16.56 \%} \\ 
 \midrule
\multirow{2}{*}{\rotatebox[origin=c]{0}{ResNet-RS 101}} & Baseline & \textbf{1.78} & \textbf{78.75 \%} & \textbf{94.21 \%} & \textbf{5.55 \si{\hour}} & \textbf{---} & --- \\
 & OIALR  & 1.81 & 77.95 \% & 93.95 \% & 5.92 \si{\hour} & 104.66 \% & \textbf{15.66 \%} \\
\bottomrule
\end{tabular}
\end{table*}
\begin{figure}[tb]
     \centering
     \begin{subfigure}[t]{\linewidth}
        \centering
        \includegraphics[width=\linewidth]{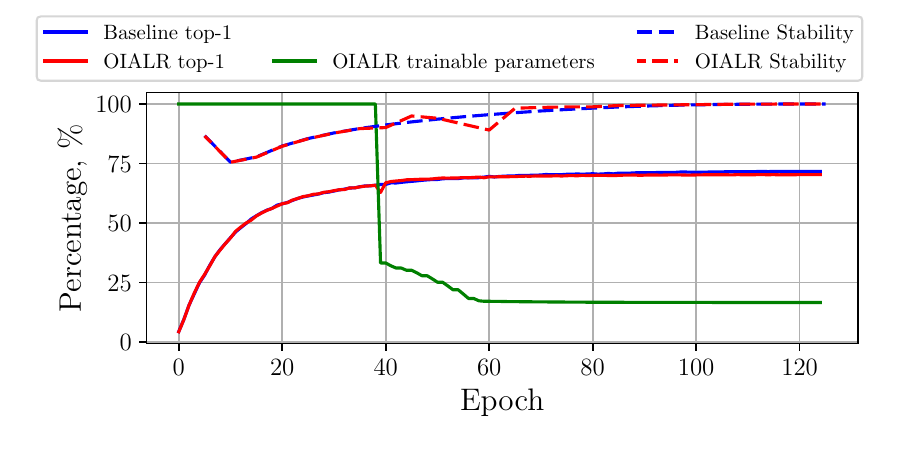}
        \caption{Top-1 validation accuracy, percentage of trainable parameters as compared to the traditional network, and average \textsl{Stability} measured with a five-epoch frequency.\\}
        \label{fig:imagenet-top1}
     \end{subfigure}
     \hfill
     \begin{subfigure}[b]{\linewidth}
         \centering
          \includegraphics[width=\linewidth]{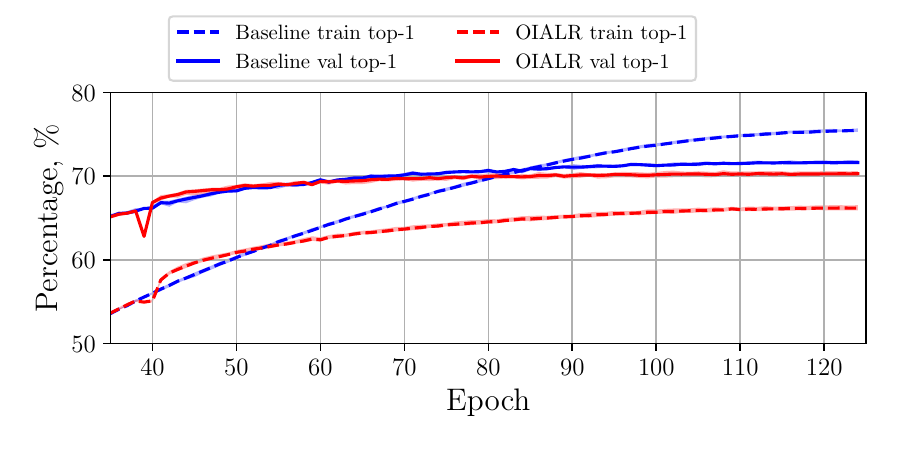}
          \caption{Top-1 accuracies for training and validation with baseline and OIALR training methods.\\}
          \label{fig:imagenet-train}
     \end{subfigure}
    \caption{Training of a ViT-B/16 network on ImageNet-2012 over 125 epochs.\vspace{0.6cm}}
    \label{fig:imagenet-curves}
\end{figure}
For the first experiment, we trained the Vision Transformer (ViT)-B/16 model~\citep{dosovitskiy2021vit} on the ImageNet-2012 dataset~\citep{beyer2020imagenet} using the ReaL validation labels~\citep{beyer2020imagenet}. 
The considerable parameter count of this model provided a rigorous test for the OIALR training method.
We maintained identical HPs for both full-rank and adaptive low-rank training. 
To reduce the environmental impact, we trained for 125 epochs instead of the original 300~\citep{dosovitskiy2021vit}. 
By this point, validation accuracy had nearly stabilized, as shown in \Cref{fig:imagenet-train}.
Additionally, we used an image resolution of $160\times 160$ instead of $224 \times 224$ to further reduce the energy consumption. 

The results are shown in \Cref{fig:imagenet-curves} and \Cref{tab:imagenet}.
\Cref{fig:imagenet-top1} presents the top-1 validation score, the percentage of trainable parameters relative to the full-rank model, and the average network \textsl{Stability} (as shown in \Cref{fig:stability}) throughout training.
Notably, the baseline \textsl{Stability} increases smoothly throughout training, while OIALR's \textsl{Stability} is less consistent due to the reductions in the weights' ranks. 
This arises from the fact that the $\boldsymbol{U}\boldsymbol{V}^T$ from five epochs prior contains more basis vectors than the current $\boldsymbol{U}\boldsymbol{V}^T$.

As evident in \Cref{fig:imagenet-train}, there is a momentary accuracy drop when the network transitions from full-rank to its $\boldsymbol{U}\boldsymbol{\Sigma}\boldsymbol{V}^T$ representation, but it swiftly rebounds, surpassing previous performance.
We theorize that this is caused by the residual momentum states in the optimizer `pushing' the network in different directions.

In this untuned case, OIALR training required approximately the same amount of time to train (~\SI{1}{\percent} longer) while the models maintained similar performance (\SI{1.34(0.39)}{\percent} decrease in top-1 accuracy) than traditional full-rank training.
OIALR reduced the number of trainable parameters to \SI{16.56(0.23)}{\percent} of the full-rank parameters.
\Cref{fig:imagenet-train} shows that the full-rank model has entered the overfitting regime, where training accuracy continues increasing while validation accuracy plateaus, whereas the low-rank model has not.

\subsection{Comparison with related low-rank and sparse training methods}
\label{sec:comp}

To show where OIALR fits into the landscape of full-to-low-rank, low-rank, and sparse training methods, we performed a comparative analysis shown in \Cref{tab:comparison}.
In these experiments, the baseline and compression methods use the same HPs. 

OIALR and DLRT~\citep{schotthofer2022lowrank} are SVD-based low-rank factorization methods.
LRNN~\citep{idelbayev2020learningrank} uses a traditional two-matrix representation ($\boldsymbol{W} \approx \boldsymbol{A}\boldsymbol{B}$).
CP~\citep{he2017cp}, SFP~\citep{he2018sfp}, PP-1~\citep{singh2019pp}, and ThiNet~\citep{luo2017thinet} are structured pruning methods which use channel or filter pruning for convolution layers.
GAL~\citep{lin2019gal} and RNP~\citep{lin2017rnp} are unstructured pruning methods; RigL~\citep{evci2020rigl} is a sparse training method.

Overall, OIALR demonstrates competitive performance across different architectures and datasets.
Despite producing subpar results on ResNet-50, it showed a marginal accuracy improvement over the baseline for VGG16~\citep{liu2015vgg16} on CIFAR-10.
We theorize that since VGG16 is known to be over-parameterized, there are more basis vectors that can be removed. 
By eliminating these less useful basis vectors, the network can focus on those remaining to train a performant low-rank network.

\begin{table}[t]
\caption{Comparison of OIALR with various compression methods. `Diff. to baseline' refers to the difference in top-1 validation (ImageNet-2012) or test (CIFAR-10) accuracy between the baseline and the listed methods. Positive values indicate that the listed method outperforms the traditionally trained network. Absence of data indicated by `---'. For non-OIALR results see \protect\citet{schotthofer2022lowrank,evci2020rigl}.}
\label{tab:comparison}
\centering
\begin{tabular}{@{}c|l c c c @{}}
\toprule
 & \begin{tabular}[c]{@{}c@{}}Training\\method \end{tabular} & \begin{tabular}[c]{@{}c@{}}Diff. to\\baseline \end{tabular} & Compression & \begin{tabular}[c]{@{}c@{}}Trainable \\ parameters \end{tabular} \\ \midrule
\multirow{7}{*}{\rotatebox[origin=c]{90}{\begin{tabular}[c]{@{}c@{}}ResNet-50 \\ ImageNet-2012 \end{tabular}}} & OIALR & -1.72 \% & 82.55 \% & 15.15 \% \\
 & DLRT & -0.56 \% & 54.10 \% & \textbf{14.20 \%}\\
 & PP-1 & \textbf{-0.20 \%} & 44.20 \% & --- \\
 & CP & -1.40 \% & 50.00 \% & ---\\
 & SFP & \textbf{-0.20 \%} & 41.80 \% & ---\\
 & ThiNet & -1.50 \% & 36.90 \% & --- \\
 & RigL & -2.20 \% & \textbf{20.00 \%} & ---\\ \midrule
\multirow{7}{*}{\rotatebox[origin=c]{90}{\hspace{4mm}\begin{tabular}[c]{@{}c@{}}VGG16 \\ ImageNet-2012 \end{tabular}}} & OIALR & -1.53 \% & \textbf{36.52 \%} & \textbf{4.77 \%} \\
 & DLRT & -2.19 \% & 86.00 \% & 78.40 \% \\
 & PP-1 & \textbf{-0.19 \%} & 80.20 \% & --- \\
 & CP & -1.80 \% & 80.00 \% & --- \\
 & ThiNet & -0.47 \% & 69.04 \% & --- \\
 & RNP(3X) & -2.43 \% & 66.67 \% & --- \\ \midrule
\multirow{7}{*}{\rotatebox[origin=c]{90}{\hspace{12mm}\begin{tabular}[c]{@{}c@{}}VGG16 \\ CIFAR-10 \end{tabular}}} &  OIALR & \textbf{0.10 \%} & \textbf{27.05 \%} & \textbf{13.88 \%} \\
 & DLRT & -1.89 \% & 56.00 \% & 77.50 \% \\
 & GAL & -1.87 \% & 77.00  \% & ---\\
 & LRNN & -1.90 \% & 60.00 \% & ---\\ \bottomrule
\end{tabular}
\end{table}

\subsection{Ablation study on mini ViT on CIFAR-10}
\label{sec:cifar}

\begin{table*}[]
\caption{A mini ViT trained on CIFAR-10. `OIALR, tuned' training runs used tuned HPs, while `OIALR' used the same HPs as the baseline. Accuracies and loss values are determined on the test dataset.}
\label{tab:cifar10}
\centering
\begin{tabular}{@{}l c c c c c c @{}}
\toprule
Training method & Loss & Top-1 accuracy & Top-5 accuracy & Time to train & Compression & Trainable parameters    \\ \midrule
Traditional & 0.88 & 85.17 \% & 98.34 \% & 12.14 \si{\min} & --- & --- \\
OIALR  & 0.91 & 83.05 \% & 98.38 \% & 11.99 \si{\min} & 146.78 \% & 30.98 \% \\
OIALR, tuned & \textbf{0.85} & \textbf{86.33 \%} & \textbf{98.53 \%} & \textbf{11.19 \si{\min}} & \hspace{2mm}\textbf{55.06 \%} & \hspace{2mm}\textbf{9.97 \%} \\
\bottomrule
\end{tabular}
\end{table*}

\begin{figure}[tb]
    \centering
    \includegraphics[width=\linewidth]{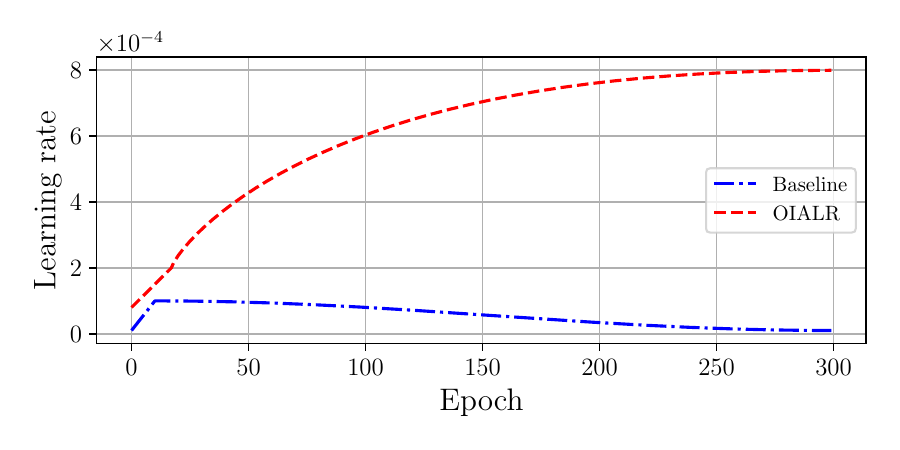}
    \caption{Learning rate schedules for baseline and OIALR training for a mini ViT on CIFAR-10. OIALR training learning rate schedule determined by HP search.\vspace{0.6cm}}
    \label{fig:cifar}
\end{figure}

To show how OIALR performs with proper tuning, we trained a reduced-size ViT model on the CIFAR-10 dataset with and without tuning.
The runs without tuning use the same HPs as the baseline runs.
As reduced-size ViT models have been shown to perform superbly~\citep{hassani2022minivit} at a fraction of the compute time, we elect to use a ViT-B/16 variant with a patch size of eight, six layers, and six attention heads in this experiment (original values are a patch size of 16, 12 layers, and 12 attention heads).
The results of this experiment are shown in \Cref{tab:cifar10} and \Cref{fig:cifar}.

Interestingly, the best learning rate schedule for the OIALR training method discovered through HP search increases the learning rate as the number of parameters decreases, see \Cref{fig:cifar}.
This result makes intuitive sense: as the number of trainable parameters decreases, the learning rate applied to the gradients of the remaining parameters can be increased without the model degrading.

Although the untuned OIALR model reduced the trainable parameters by \SI{69.02(0.07)}{\percent}, the top-1 test accuracy dropped by over \SI{2}{\percent}.
In contrast, the tuned OIALR model reduced the number of trainable parameters by \SI{90.03(0.13)}{\percent} while increasing predictive performance over the baseline from \SI{85.17(0.42)}{\percent} to \SI{86.33(0.71)}{\percent}.
Furthermore, training time was reduced by \SI{8.52(0.82)}{\percent} over the baseline.

\subsection{Ablation study on Autoformer on ETTm2}
\label{sec:ettm2}

\begin{table*}[]
\caption{Training of the Autoformer model on the ETTm2 dataset. Baseline and untuned OIALR HPs were the default parameters from \protect\citet{wu2021autoformer}. Tuned OIALR HPs were found via \texttt{Propulate}. Prediction lengths (PL) in the leftmost column are in \SI{15}{\min} time steps. The optimal value for mean squared error (MSE) and mean absolute error (MAE) is zero.}
\label{tab:ettm2}
\centering
\begin{tabular}{@{}r|l c c c c@{}}
\toprule
PL & Training method & MSE & MAE & Compression & Trainable parameters \\ \midrule
\multirow{3}{*}{96} & Baseline &  0.2145 & 0.2994 & --- & --- \\
                    & OIALR & 0.2140 & 0.2974 & 182.06 \% & 46.16 \% \\
                    & OIALR, tuned & \textbf{0.2112} & \textbf{0.2942} & \hspace{2mm}\textbf{47.84 \%} & \textbf{12.19} \% \\ \midrule
\multirow{3}{*}{192} & Baseline     & 0.2737 & 0.3356 & \textbf{---} & --- \\
                     & OIALR        & 0.2773 & 0.3336 & 163.35 \% & 105.31 \% \\
                     & OIALR, tuned & \textbf{0.2686} & \textbf{0.3305} & 105.31 \% & \hspace{2mm}\textbf{27.15 \%} \\ \midrule
\multirow{3}{*}{336} & Baseline     & 0.3277 & 0.3640 & --- & --- \\
                     & OIALR        & 0.3253 & 0.3863 & 179.87 \% & 45.67 \% \\
                     & OIALR, tuned & \textbf{0.3212} & \textbf{0.3591} & \hspace{2mm}\textbf{27.30 \%} & \hspace{2mm}\textbf{7.14 \%} \\ \midrule
\multirow{3}{*}{720} & Baseline     & 0.4194 & 0.4157 & --- & --- \\
                     & OIALR        & 0.4213 & 0.4186 & 194.13 \% & 51.33 \% \\
                     & OIALR, tuned & \textbf{0.4120} & \textbf{0.4147} & \hspace{2mm}\textbf{13.55 \%} & \hspace{2mm}\textbf{4.46 \%} \\
\bottomrule
\end{tabular}
\end{table*}

\begin{figure}[t]
     \centering
     \begin{subfigure}[b]{\linewidth}
        \centering
        \includegraphics[width=\linewidth]{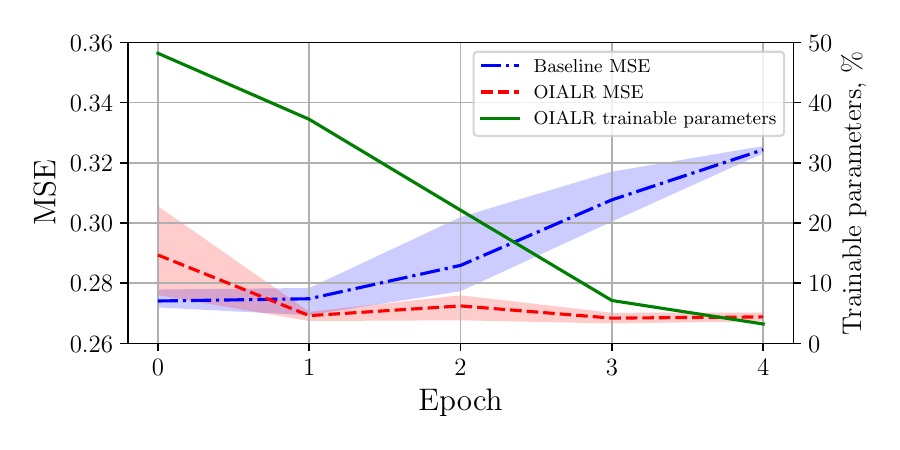}
        \caption{Prediction length 192\\~}
        \label{fig:af-192}
     \end{subfigure}
     \hfill
     \begin{subfigure}[b]{\linewidth}
         \centering
          \includegraphics[width=\linewidth]{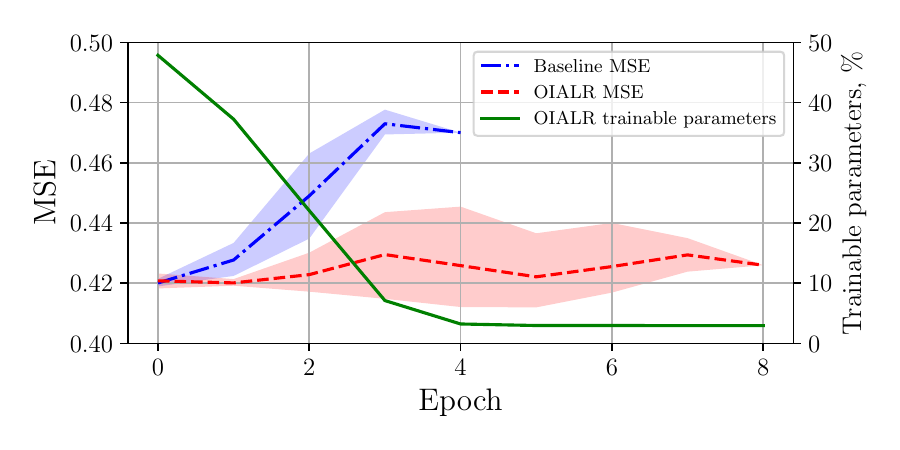}
          \caption{Prediction length 720\\~}
          \label{fig:af-720}
     \end{subfigure}
    \caption{MSE and the percentage of trainable parameters relative to the full-rank model for the Autoformer trained on the ETTm2 dataset using two different prediction lengths in \SI{15}{\min} time steps.\\~}
    \label{fig:autoformer}
    
\end{figure}

This use case serves as a crucial test for the OIALR method, showcasing its versatility by applying it to a model in a radically different domain. 
Furthermore, it validates that the findings depicted in \Cref{fig:stability} remain applicable in non-image scenarios.

The Electricity Transformer Dataset~\citep{zhou2021informer} (ETT) measures load and oil temperature of electrical transformers.
It contains 70,000 measurements at various levels of granularity, each with seven features, and is primarily used for time series forecasting.
We focus on the ETTm2 dataset, which has a 15-minute resolution.
Common prediction lengths for this dataset are 96, 192, 336, and 720 time steps.
The Autoformer~\citep{wu2021autoformer}, a well-known transformer model in Hugging Face's repository, differs from the other tested transformers by using auto-correlation layers and one-dimensional convolutions.
Due to its success, it was deployed at the 2022 Winter Olympics for weather forecasting.

Given the baseline's susceptibility to overfitting (see \Cref{fig:autoformer}), we initiate training in low rank instead of transitioning during training.
Although some overfitting is observed in the OIALR results, it is considerably less severe than in the baseline.
As \Cref{tab:ettm2} indicates, the tuned OIALR models were more accurate across all prediction lengths with a drastically decreased number of parameters. 

The untuned OIALR models outperformed the baseline in some cases and succeeded in reducing the number of trainable parameters to \SI{45.72}{\percent} on average.
As explained in \Cref{sec:oialr}, these models required more parameters than the baseline model due to the shapes of the model's $\boldsymbol{U}\boldsymbol{\Sigma}\boldsymbol{V}^T$ representation.
The tuned OIALR measurements generally showed a much more successful compression percentage.
Interestingly, the tuned OIALR model required more trainable parameters for predicting shorter time spans. 

In contrast to the previous experiment, the best learning rate scheduler found for this use case more closely resembles a traditional scheduler, featuring a warm-up phase followed by a gradual decay.
This may be related to the fact that the networks, both low-rank and full-rank, overfit the training dataset quickly.

\section{Conclusion}
There has long been curiosity about how a neural network learns.
This study aimed to shed light on this question by exploring the nature of neural network weights during training through their singular value decomposition.
Our findings revealed that the orthogonal component of a neural network's weights stabilizes early in the training process.
Building on this discovery, we introduced Orthogonality-Informed Adaptive Low-Rank (OIALR) training. 

We evaluated OIALR by training low-rank versions of widely used and state-of-the-art neural networks across diverse data modalities and tasks.
OIALR demonstrated superior performance when compared against other low-rank methods, even when employing the default hyperparameter settings used in traditional model training.
While our approach may not directly surpass traditional training techniques in all scenarios, it can outperform them in terms of both accuracy and training time when tuned appropriately.

OIALR's true strength lies in substantially reducing the number of trainable parameters of the final model, thereby facilitating model fine-tuning, transfer learning, and deployment on resource-constrained devices. 
This reduction also contributes to reducing the data transfer requirements during distributed training, reducing the gap between expensive, top-tier clusters and more affordable options.

Integrating orthogonality-based training methods into the deep learning researcher's toolkit offers promising possibilities for a wide range of applications. 
With this work, we hope to inspire further exploration and refinement of orthogonality-informed methods, ultimately advancing the field of machine learning and its practicality across diverse domains.





\begin{ack}
This work was performed on the HoreKa supercomputer funded by the Ministry of Science, Research and the Arts Baden-W\"urttemberg and by the Federal Ministry of Education and Research.
This work is supported by the Helmholtz Association Initiative and Networking Fund under the Helmholtz AI platform grant and the HAICORE@KIT partition.
\end{ack}



\bibliography{references}

\begin{thebibliography}{47}
\providecommand{\natexlab}[1]{#1}
\providecommand{\url}[1]{\texttt{#1}}
\expandafter\ifx\csname urlstyle\endcsname\relax
  \providecommand{\doi}[1]{doi: #1}\else
  \providecommand{\doi}{doi: \begingroup \urlstyle{rm}\Url}\fi

\bibitem[Bejani and Ghatee(2020)]{bejani2020lowrank-regularize}
M.~M. Bejani and M.~Ghatee.
\newblock Adaptive {Low}-{Rank} {Factorization} to regularize shallow and deep neural networks, May 2020.
\newblock URL \url{http://arxiv.org/abs/2005.01995}.
\newblock arXiv:2005.01995 [cs, stat].

\bibitem[Bello et~al.(2021)Bello, Fedus, Du, et~al.]{bello2021resnetrs}
I.~Bello, W.~Fedus, X.~Du, et~al.
\newblock Revisiting {ResNets}: {Improved} {Training} and {Scaling} {Strategies}, Mar. 2021.
\newblock URL \url{http://arxiv.org/abs/2103.07579}.
\newblock arXiv:2103.07579 [cs].

\bibitem[Beyer et~al.(2020)Beyer, Hénaff, Kolesnikov, et~al.]{beyer2020imagenet}
L.~Beyer, O.~J. Hénaff, A.~Kolesnikov, et~al.
\newblock Are we done with {ImageNet}?, June 2020.
\newblock URL \url{http://arxiv.org/abs/2006.07159}.
\newblock arXiv:2006.07159 [cs].

\bibitem[Cahyawijaya et~al.(2021)Cahyawijaya, Winata, Lovenia, et~al.]{cahyawijaya2021greenformer}
S.~Cahyawijaya, G.~I. Winata, H.~Lovenia, et~al.
\newblock Greenformer: {Factorization} {Toolkit} for {Efficient} {Deep} {Neural} {Networks}, Oct. 2021.
\newblock URL \url{http://arxiv.org/abs/2109.06762}.
\newblock arXiv:2109.06762 [cs].

\bibitem[Ceruti et~al.(2021)Ceruti, Kusch, and Lubich]{ceruti2021rank-adaptive}
G.~Ceruti, J.~Kusch, and C.~Lubich.
\newblock A rank-adaptive robust integrator for dynamical low-rank approximation, Apr. 2021.
\newblock URL \url{http://arxiv.org/abs/2104.05247}.
\newblock arXiv:2104.05247 [cs, math].

\bibitem[Coquelin et~al.(2021)Coquelin, Sedona, Riedel, and Götz]{coquelin2021rsapplication}
D.~Coquelin, R.~Sedona, M.~Riedel, and M.~Götz.
\newblock Evolutionary {Optimization} of {Neural} {Architectures} in {Remote} {Sensing} {Classification} {Problems}.
\newblock In \emph{2021 {IEEE} {International} {Geoscience} and {Remote} {Sensing} {Symposium} {IGARSS}}, pages 1587--1590, July 2021.
\newblock \doi{10.1109/IGARSS47720.2021.9554309}.
\newblock ISSN: 2153-7003.

\bibitem[Deng et~al.(2020)Deng, Li, Han, Shi, and Xie]{deng2020compressionsurevy}
L.~Deng, G.~Li, S.~Han, L.~Shi, and Y.~Xie.
\newblock Model {Compression} and {Hardware} {Acceleration} for {Neural} {Networks}: {A} {Comprehensive} {Survey}.
\newblock \emph{Proceedings of the IEEE}, 108\penalty0 (4):\penalty0 485--532, Apr. 2020.
\newblock ISSN 1558-2256.
\newblock \doi{10.1109/JPROC.2020.2976475}.
\newblock Conference Name: Proceedings of the IEEE.

\bibitem[Dosovitskiy et~al.(2021)Dosovitskiy, Beyer, Kolesnikov, et~al.]{dosovitskiy2021vit}
A.~Dosovitskiy, L.~Beyer, A.~Kolesnikov, et~al.
\newblock An {Image} is {Worth} 16x16 {Words}: {Transformers} for {Image} {Recognition} at {Scale}, June 2021.
\newblock URL \url{http://arxiv.org/abs/2010.11929}.
\newblock arXiv:2010.11929 [cs] version: 2.

\bibitem[Evci et~al.(2020)Evci, Gale, Menick, et~al.]{evci2020rigl}
U.~Evci, T.~Gale, J.~Menick, et~al.
\newblock {Rigging the Lottery: Making All Tickets Winners}.
\newblock In \emph{Proceedings of the 37th International Conference on Machine Learning}, ICML'20. JMLR.org, 2020.

\bibitem[Gotmare et~al.(2019)Gotmare, Keskar, Xiong, and Socher]{gotmare2018lrwarmup}
A.~Gotmare, N.~S. Keskar, C.~Xiong, and R.~Socher.
\newblock {A Closer Look at Deep Learning Heuristics: Learning rate restarts, Warmup and Distillation}.
\newblock In \emph{International Conference on Learning Representations}, 2019.
\newblock URL \url{https://openreview.net/forum?id=r14EOsCqKX}.

\bibitem[Guo et~al.(2023)Guo, Yolwas, and Slamu]{guo2023conformerlowrank}
T.~Guo, N.~Yolwas, and W.~Slamu.
\newblock Efficient {Conformer} for {Agglutinative} {Language} {ASR} {Model} {Using} {Low}-{Rank} {Approximation} and {Balanced} {Softmax}.
\newblock \emph{Applied Sciences}, 13\penalty0 (7):\penalty0 4642, Jan. 2023.
\newblock ISSN 2076-3417.
\newblock \doi{10.3390/app13074642}.
\newblock URL \url{https://www.mdpi.com/2076-3417/13/7/4642}.
\newblock Number: 7 Publisher: Multidisciplinary Digital Publishing Institute.

\bibitem[Hassani et~al.(2022)Hassani, Walton, Shah, et~al.]{hassani2022minivit}
A.~Hassani, S.~Walton, N.~Shah, et~al.
\newblock Escaping the {Big} {Data} {Paradigm} with {Compact} {Transformers}, June 2022.
\newblock URL \url{http://arxiv.org/abs/2104.05704}.
\newblock arXiv:2104.05704 [cs] version: 4.

\bibitem[He et~al.(2017)He, Zhang, and Sun]{he2017cp}
Y.~He, X.~Zhang, and J.~Sun.
\newblock {Channel Pruning for Accelerating Very Deep Neural Networks}.
\newblock In \emph{2017 IEEE International Conference on Computer Vision (ICCV)}, pages 1398--1406, 2017.
\newblock \doi{10.1109/ICCV.2017.155}.

\bibitem[He et~al.(2018)He, Kang, Dong, et~al.]{he2018sfp}
Y.~He, G.~Kang, X.~Dong, et~al.
\newblock {Soft Filter Pruning for Accelerating Deep Convolutional Neural Networks}.
\newblock In \emph{Proceedings of the 27th International Joint Conference on Artificial Intelligence}, IJCAI'18, page 2234–2240. AAAI Press, 2018.
\newblock ISBN 9780999241127.

\bibitem[Hssayni et~al.(2022)Hssayni, Joudar, and Ettaouil]{hssayni2022_krr-cnn}
E.~h. Hssayni, N.-E. Joudar, and M.~Ettaouil.
\newblock {KRR}-{CNN}: kernels redundancy reduction in convolutional neural networks.
\newblock \emph{Neural Computing and Applications}, 34\penalty0 (3):\penalty0 2443--2454, Feb. 2022.
\newblock ISSN 1433-3058.
\newblock \doi{10.1007/s00521-021-06540-3}.
\newblock URL \url{https://doi.org/10.1007/s00521-021-06540-3}.

\bibitem[Hsu et~al.(2022)Hsu, Hua, Chang, et~al.]{hsu2022svdllm}
Y.-C. Hsu, T.~Hua, S.~Chang, et~al.
\newblock {Language model compression with weighted low-rank factorization}.
\newblock In \emph{International Conference on Learning Representations}, 2022.
\newblock URL \url{https://openreview.net/forum?id=uPv9Y3gmAI5}.

\bibitem[Hu et~al.(2022)Hu, Shen, Wallis, et~al.]{hu2022lora}
E.~J. Hu, Y.~Shen, P.~Wallis, et~al.
\newblock Lo{RA}: Low-rank adaptation of large language models.
\newblock In \emph{International Conference on Learning Representations}, 2022.
\newblock URL \url{https://openreview.net/forum?id=nZeVKeeFYf9}.

\bibitem[Idelbayev and Carreira-Perpinan(2020)]{idelbayev2020learningrank}
Y.~Idelbayev and M.~A. Carreira-Perpinan.
\newblock Low-{Rank} {Compression} of {Neural} {Nets}: {Learning} the {Rank} of {Each} {Layer}.
\newblock In \emph{2020 {IEEE}/{CVF} {Conference} on {Computer} {Vision} and {Pattern} {Recognition} ({CVPR})}, pages 8046--8056, June 2020.
\newblock \doi{10.1109/CVPR42600.2020.00807}.
\newblock ISSN: 2575-7075.

\bibitem[Ji and Telgarsky(2020)]{ji2020directional}
Z.~Ji and M.~Telgarsky.
\newblock Directional convergence and alignment in deep learning.
\newblock In H.~Larochelle, M.~Ranzato, R.~Hadsell, M.~Balcan, and H.~Lin, editors, \emph{Advances in Neural Information Processing Systems}, volume~33, pages 17176--17186. Curran Associates, Inc., 2020.
\newblock URL \url{https://proceedings.neurips.cc/paper_files/paper/2020/file/c76e4b2fa54f8506719a5c0dc14c2eb9-Paper.pdf}.

\bibitem[L. and H.(2018)]{loshchilov2018adamw}
I.~L. and F.~H.
\newblock {Fixing Weight Decay Regularization in Adam}, 2018.
\newblock URL \url{https://openreview.net/forum?id=rk6qdGgCZ}.

\bibitem[Lin et~al.(2017)Lin, Rao, Lu, and Zhou]{lin2017rnp}
J.~Lin, Y.~Rao, J.~Lu, and J.~Zhou.
\newblock {Runtime Neural Pruning}.
\newblock In I.~Guyon, U.~V. Luxburg, S.~Bengio, H.~Wallach, R.~Fergus, S.~Vishwanathan, and R.~Garnett, editors, \emph{Advances in Neural Information Processing Systems}, volume~30. Curran Associates, Inc., 2017.

\bibitem[Lin et~al.(2019)Lin, Ji, Yan, et~al.]{lin2019gal}
S.~Lin, R.~Ji, C.~Yan, et~al.
\newblock {Towards Optimal Structured CNN Pruning via Generative Adversarial Learning}.
\newblock In \emph{2019 IEEE/CVF Conference on Computer Vision and Pattern Recognition (CVPR)}, pages 2785--2794, 2019.
\newblock \doi{10.1109/CVPR.2019.00290}.

\bibitem[Liu and Deng(2015)]{liu2015vgg16}
S.~Liu and W.~Deng.
\newblock Very deep convolutional neural network based image classification using small training sample size.
\newblock In \emph{2015 3rd IAPR Asian Conference on Pattern Recognition (ACPR)}, pages 730--734, 2015.
\newblock \doi{10.1109/ACPR.2015.7486599}.

\bibitem[Loshchilov and Hutter(2017)]{loshchilov2017cosine}
I.~Loshchilov and F.~Hutter.
\newblock {SGDR}: {Stochastic} {Gradient} {Descent} with {Warm} {Restarts}, May 2017.
\newblock URL \url{http://arxiv.org/abs/1608.03983}.
\newblock arXiv:1608.03983 [cs, math].

\bibitem[Luo et~al.(2017)Luo, Wu, and Lin]{luo2017thinet}
J.-H. Luo, J.~Wu, and W.~Lin.
\newblock {ThiNet: A Filter Level Pruning Method for Deep Neural Network Compression}.
\newblock In \emph{2017 IEEE International Conference on Computer Vision (ICCV)}, pages 5068--5076, 2017.
\newblock \doi{10.1109/ICCV.2017.541}.

\bibitem[Mahabadi et~al.(2021)Mahabadi, Henderson, and Ruder]{mahabadi2021compacter}
R.~K. Mahabadi, J.~Henderson, and S.~Ruder.
\newblock Compacter: Efficient low-rank hypercomplex adapter layers.
\newblock In A.~Beygelzimer, Y.~Dauphin, P.~Liang, and J.~W. Vaughan, editors, \emph{Advances in Neural Information Processing Systems}, 2021.
\newblock URL \url{https://openreview.net/forum?id=bqGK5PyI6-N}.

\bibitem[Nitish et~al.(2014)Nitish, Hinton, Krizhevsky, et~al.]{srivastava2014dropout}
S.~Nitish, G.~Hinton, A.~Krizhevsky, et~al.
\newblock {Dropout: A Simple Way to Prevent Neural Networks from Overfitting}.
\newblock \emph{Journal of Machine Learning Research}, 15\penalty0 (56):\penalty0 1929--1958, 2014.
\newblock URL \url{http://jmlr.org/papers/v15/srivastava14a.html}.

\bibitem[Paszke et~al.(2019)Paszke, Gross, Massa, et~al.]{paszke2018pytorch}
A.~Paszke, S.~Gross, F.~Massa, et~al.
\newblock Pytorch: An imperative style, high-performance deep learning library.
\newblock In \emph{Advances in Neural Information Processing Systems}, volume~32. Curran Associates, Inc., 2019.

\bibitem[Phan et~al.(2020)Phan, Sobolev, Sozykin, et~al.]{phan2020stablecnn}
A.-H. Phan, K.~Sobolev, K.~Sozykin, et~al.
\newblock Stable {Low}-{Rank} {Tensor} {Decomposition} for {Compression} of {Convolutional} {Neural} {Network}.
\newblock In A.~Vedaldi, H.~Bischof, T.~Brox, and J.-M. Frahm, editors, \emph{Computer {Vision} – {ECCV} 2020}, Lecture {Notes} in {Computer} {Science}, pages 522--539, Cham, 2020. Springer International Publishing.
\newblock ISBN 978-3-030-58526-6.
\newblock \doi{10.1007/978-3-030-58526-6_31}.

\bibitem[Povey et~al.(2018)Povey, Cheng, Wang, et~al.]{povey2018semiorthogonal}
D.~Povey, G.~Cheng, Y.~Wang, et~al.
\newblock Semi-{Orthogonal} {Low}-{Rank} {Matrix} {Factorization} for {Deep} {Neural} {Networks}.
\newblock In \emph{Interspeech 2018}, pages 3743--3747. ISCA, Sept. 2018.
\newblock \doi{10.21437/Interspeech.2018-1417}.
\newblock URL \url{https://www.isca-speech.org/archive/interspeech_2018/povey18_interspeech.html}.

\bibitem[Psichogios and Ungar(1994)]{psichogios1994svd-net}
D.~Psichogios and L.~Ungar.
\newblock {SVD}-{NET}: an algorithm that automatically selects network structure.
\newblock \emph{IEEE Transactions on Neural Networks}, 5\penalty0 (3):\penalty0 513--515, May 1994.
\newblock ISSN 1941-0093.
\newblock \doi{10.1109/72.286929}.
\newblock Conference Name: IEEE Transactions on Neural Networks.

\bibitem[Russakovsky et~al.(2015)Russakovsky, Deng, Su, et~al.]{russakovsky2015imagenet}
O.~Russakovsky, J.~Deng, H.~Su, et~al.
\newblock {ImageNet} {Large} {Scale} {Visual} {Recognition} {Challenge}.
\newblock \emph{International Journal of Computer Vision}, 115\penalty0 (3):\penalty0 211--252, Dec. 2015.
\newblock ISSN 1573-1405.
\newblock \doi{10.1007/s11263-015-0816-y}.
\newblock URL \url{https://doi.org/10.1007/s11263-015-0816-y}.

\bibitem[Schotthöfer et~al.(2022)Schotthöfer, Zangrando, Kusch, et~al.]{schotthofer2022lowrank}
S.~Schotthöfer, E.~Zangrando, J.~Kusch, et~al.
\newblock Low-rank lottery tickets: finding efficient low-rank neural networks via matrix differential equations.
\newblock \emph{Advances in Neural Information Processing Systems}, 35:\penalty0 20051--20063, Dec. 2022.
\newblock URL \url{https://proceedings.neurips.cc/paper_files/paper/2022/hash/7e98b00eeafcdaeb0c5661fb9355be3a-Abstract-Conference.html}.

\bibitem[Singh et~al.(2019)Singh, Verma, Rai, and Namboodiri]{singh2019pp}
P.~Singh, V.~K. Verma, P.~Rai, and V.~P. Namboodiri.
\newblock {Play and Prune: Adaptive Filter Pruning for Deep Model Compression}.
\newblock In \emph{Proceedings of the 28th International Joint Conference on Artificial Intelligence}, IJCAI'19, page 3460–3466. AAAI Press, 2019.
\newblock ISBN 9780999241141.

\bibitem[Taubert et~al.(2023)Taubert, Weiel, Coquelin, et~al.]{taubert2023massively}
O.~Taubert, M.~Weiel, D.~Coquelin, et~al.
\newblock Massively parallel genetic optimization through asynchronous propagation of populations.
\newblock In \emph{International Conference on High Performance Computing}, pages 106--124. Springer, 2023.

\bibitem[Touvron et~al.(2021)Touvron, Cord, Douze, et~al.]{touvron2021augments}
H.~Touvron, M.~Cord, M.~Douze, et~al.
\newblock {Training data-efficient image transformers \& distillation through attention}.
\newblock In M.~Meila and T.~Zhang, editors, \emph{Proceedings of the 38th International Conference on Machine Learning}, volume 139 of \emph{Proceedings of Machine Learning Research}, pages 10347--10357. PMLR, 18--24 Jul 2021.
\newblock URL \url{https://proceedings.mlr.press/v139/touvron21a.html}.

\bibitem[Waleffe and Rekatsinas(2020)]{waleffe2020pcanets}
R.~Waleffe and T.~Rekatsinas.
\newblock {Principal Component Networks: Parameter Reduction Early in Training}.
\newblock \emph{CoRR}, abs/2006.13347, 2020.
\newblock URL \url{https://arxiv.org/abs/2006.13347}.

\bibitem[Wang et~al.(2020)Wang, Wohlwend, and Lei]{wang2020structured}
Z.~Wang, J.~Wohlwend, and T.~Lei.
\newblock Structured pruning of large language models.
\newblock In B.~Webber, T.~Cohn, Y.~He, and Y.~Liu, editors, \emph{Proceedings of the 2020 Conference on Empirical Methods in Natural Language Processing (EMNLP)}, pages 6151--6162, Online, Nov. 2020. Association for Computational Linguistics.
\newblock \doi{10.18653/v1/2020.emnlp-main.496}.
\newblock URL \url{https://aclanthology.org/2020.emnlp-main.496}.

\bibitem[Wightman et~al.(2023)Wightman, Raw, Soare, et~al.]{timm_2023}
R.~Wightman, N.~Raw, A.~Soare, et~al.
\newblock rwightman/pytorch-image-models: v0.8.10dev0 {Release}, Feb. 2023.
\newblock URL \url{https://zenodo.org/record/4414861}.

\bibitem[Winata et~al.(2020)Winata, Cahyawijaya, Lin, Liu, and Fung]{winata2020lowranktrans-speech}
G.~I. Winata, S.~Cahyawijaya, Z.~Lin, Z.~Liu, and P.~Fung.
\newblock Lightweight and {Efficient} {End}-{To}-{End} {Speech} {Recognition} {Using} {Low}-{Rank} {Transformer}.
\newblock In \emph{{ICASSP} 2020 - 2020 {IEEE} {International} {Conference} on {Acoustics}, {Speech} and {Signal} {Processing} ({ICASSP})}, pages 6144--6148, May 2020.
\newblock \doi{10.1109/ICASSP40776.2020.9053878}.
\newblock ISSN: 2379-190X.

\bibitem[Wu et~al.(2021)Wu, Xu, Wang, and Long]{wu2021autoformer}
H.~Wu, J.~Xu, J.~Wang, and M.~Long.
\newblock {Autoformer: Decomposition Transformers with Auto-Correlation for Long-Term Series Forecasting}.
\newblock In \emph{Advances in Neural Information Processing Systems}, 2021.

\bibitem[Xu and McAuley(2023)]{xu2023compresssurvey}
C.~Xu and J.~McAuley.
\newblock A {Survey} on {Model} {Compression} and {Acceleration} for {Pretrained} {Language} {Models}.
\newblock \emph{Proceedings of the AAAI Conference on Artificial Intelligence}, 37\penalty0 (9):\penalty0 10566--10575, June 2023.
\newblock ISSN 2374-3468.
\newblock \doi{10.1609/aaai.v37i9.26255}.
\newblock URL \url{https://ojs.aaai.org/index.php/AAAI/article/view/26255}.
\newblock Number: 9.

\bibitem[Xu et~al.(2019)Xu, Li, Zhang, Wen, Wang, Dai, Qi, Chen, Lin, and Xiong]{xu2019train}
Y.~Xu, Y.~Li, S.~Zhang, W.~Wen, B.~Wang, W.~Dai, Y.~Qi, Y.~Chen, W.~Lin, and H.~Xiong.
\newblock {Trained Rank Pruning for Efficient Deep Neural Networks}.
\newblock In \emph{2019 Fifth Workshop on Energy Efficient Machine Learning and Cognitive Computing - NeurIPS Edition (EMC2-NIPS)}, pages 14--17, 2019.
\newblock \doi{10.1109/EMC2-NIPS53020.2019.00011}.

\bibitem[Yang et~al.(2021)Yang, Zhang, and Sudjianto]{yang2021exnn}
Z.~Yang, A.~Zhang, and A.~Sudjianto.
\newblock Enhancing explainability of neural networks through architecture constraints.
\newblock \emph{IEEE Transactions on Neural Networks and Learning Systems}, 32\penalty0 (6):\penalty0 2610--2621, 2021.
\newblock \doi{10.1109/TNNLS.2020.3007259}.

\bibitem[Zhang et~al.(2023)Zhang, Zheng, Zhou, and Lu]{zhang2023bort}
B.~Zhang, W.~Zheng, J.~Zhou, and J.~Lu.
\newblock {Bort: Towards Explainable Neural Networks with Bounded Orthogonal Constraint}.
\newblock In \emph{The Eleventh International Conference on Learning Representations}, 2023.
\newblock URL \url{https://openreview.net/forum?id=My57qBufZWs}.

\bibitem[Zhang and Li(2022)]{zhang2022kdecay}
T.~Zhang and W.~Li.
\newblock {kDecay}: {Just} adding k-decay items on {Learning}-{Rate} {Schedule} to improve {Neural} {Networks}, Mar. 2022.
\newblock URL \url{http://arxiv.org/abs/2004.05909}.
\newblock arXiv:2004.05909 [cs].

\bibitem[Zhou et~al.(2021)Zhou, Zhang, Peng, et~al.]{zhou2021informer}
H.~Zhou, S.~Zhang, J.~Peng, et~al.
\newblock Informer: {Beyond} {Efficient} {Transformer} for {Long} {Sequence} {Time}-{Series} {Forecasting}, Mar. 2021.
\newblock URL \url{http://arxiv.org/abs/2012.07436}.
\newblock arXiv:2012.07436 [cs].

\end{thebibliography}

\appendix 

\section{Experiment Hyperparameters}
\label{sec:hypers}

Parameters not listed use the default values in the respective implementations.

\subsection{ImageNet-2012}
\label{sec:app-imagenet}
The non-default hyperparameters for all experiments on the ImageNet-2012 dataset are shown in \Cref{tab:imagenet-hps}. 
We utilized the ViT implementation from Torchvision~\cite{paszke2018pytorch} and the ResNet-RS 101 implementation from \cite{timm_2023}.

\begin{table*}[h]
\caption{Hyperparameters for training networks on ImageNet-2012. Dataset parameters are referring to the dataset transforms provided by \protect\cite{timm_2023}. LR k-decay is a parameter of the cosine learning rate decay \protect\cite{zhang2022kdecay}}
\label{tab:imagenet-hps}
\centering
\begin{tabular}{@{}llll@{}}
\toprule
\multicolumn{4}{c}{General Training Hyperparameters} \\ \midrule
Local batch size & 128 & \multicolumn{2}{c}{Learning Rate Scheduler} \\ \cmidrule(l){3-4} 
Global batch size & 1024 & Learning rate (LR) & 0.001 \\
Autocast to bfloat16 & True & Minimum learning rate & 0.00001 \\
Epochs & 125 & Warmup LR & 0.00001 \\
Label smoothing & 0.1 & LR k-decay & 1 \\
Optimizer & AdamW & Warmup epochs & 10 \\
Sync batchnorm & True &  &  \\ \midrule
\multicolumn{4}{c}{General dataset hyperparameters} \\ \midrule
Interpolation & random & Auto augment & rand-m15-mstd0.5-inc1 \\
Random erasing probability & 0.25 & crop pct & 0.9 \\
Random erasing mode & pixel & scale & (0.08, 1) \\
 &  & Training crop size & 160 \\ \midrule
\multicolumn{4}{c}{OIALR hyperparameters} \\ \midrule
Full rank first layer & False & Delay & 25000 \\
Stability frequency & 1000 & Full rank last layer & True \\
Sigma cutoff fraction & 0.1 &  &  \\ \midrule
\multicolumn{4}{c}{ResNet-RS 101 hyperparameters} \\ \midrule
Dropout & 0.25 & Validate crop size & 224 \\ \midrule
\multicolumn{4}{c}{ViT B/16 hyperparameters} \\ \midrule
Dropout & 0.1 & Hidden dim & 768 \\
Mlp dim & 3072 & Num layers & 12 \\
Num heads & 12 & Patch size & 16 \\ \bottomrule
\end{tabular}
\end{table*}




\subsection{Mini-ViT on CIFAR-10}

For training the mini-ViT we used most of the same parameters as listed in \Cref{tab:imagenet-hps} except a lower learning rate.
The utilized ViT for these experiments was from \cite{timm_2023}.
The training parameters for these experiments are shown in \Cref{tab:cifar-hps}.
The search space for \texttt{Propulate} and the parameters for the search itself are shown in \Cref{tab:propulate-cifar10-hps}

\begin{table*}[h]
\caption{Hyperparameters used for CIFAR-10 training runs. General hyperparameters used for all runs, OAILR hyperparameters use for all OAILR runs. Dataset parameters refer to implementation options in \texttt{timm} \protect\cite{timm_2023}}
\label{tab:cifar-hps}
\centering
\begin{tabular}{@{}llll@{}}
\toprule
\multicolumn{4}{c}{General Hyperparamters} \\ \midrule
Train crop size & 32 & Label smoothing & 0.1 \\
Local batch size & 256 & Optimizer & AdamW \\
Global batch size & 1024 & auto\_augment & rand-m9-mstd0.5-inc1 \\
Autocast to bfloat16 & True & Crop percent & 1 \\
Random erasing probability & 0.25 & Image scale & (0.8, 1.0) \\
Random erasing mode & pixel & Interpolation & random \\
ViT depth & 6 & ViT num heads & 6 \\
ViT qkv\_bias & False & ViT patch\_size & 8 \\
ViT embed\_dim & 768 & ViT drop\_path\_rate & 0.2 \\
ViT mlp\_ratio & 4 &  &  \\ \midrule
\multicolumn{2}{c}{Baseline} & \multicolumn{2}{c}{Tuned} \\ \midrule
LR & 0.0001 & LR & 0.0002 \\
Minimum LR & 0.00001 & Minimum LR & 0.0008 \\
Warmup LR & 0.00001 & Warmup LR & 0.00008 \\
LR k-decay & 1 & LR k-decay & 0.4 \\
Warmup epochs & 10 & Warmup epochs & 17 \\ \midrule
\multicolumn{4}{c}{OIALR hyperparameters} \\ \midrule
Delay & 4000 & Stability frequency & 1000 \\
Full rank last layer & True & Sigma cutoff fraction & 0.2 \\
Full rank first layer & False &  &  \\ \bottomrule
\end{tabular}
\end{table*}

\begin{table*}[h]
\caption{Propulate search parameters for the mini ViT on CIFAR-10}
\label{tab:propulate-cifar10-hps}
\centering
\begin{tabular}{llll}
\hline
Parameters to search over & Search space & Propulate parameter & Value \\ \hline
LR & (5e-5, 1e-3) & Crossover probability & 0.7 \\
Minimum LR & (5e-6, 1e-3) & Mutation probability & 0.4 \\
Warmup LR & (5e-6, 2e-4) & Random init probability & 0.1 \\
LR k-decay & (0.1, 2) & Number of islands & 8 \\
Warmup epochs & (1, 20) & Migration probabilty & 0.9 \\
OIALR sigma cutoff fraction & (0.01, 0.9) &  &  \\
OIALR stability frequency & (200, 1000) &  &  \\
OIALR delay & (10e2 10e3) &  &  \\ \hline
\end{tabular}
\end{table*}

\subsection{AutoFormer on ETTm2}

The learning rate schedule used in the original AutoFormer~\cite{wu2021autoformer} is a step-based schedule with fixed steps, it is denoted as `type1.'
The hyperparameters used in our experiments are listed in \Cref{tab:af-hps}.
The parameters for the hyperparameter search are listed in \Cref{tab:af-prop}.

\begin{table*}[h]
\caption{Hyperparameters used for training AutoFormer models on the ETTm2 dataset.}
\label{tab:af-hps}
\centering
\begin{tabular}{@{}llll@{}}
\toprule
\multicolumn{4}{c}{General Hyperparameters} \\ \midrule
Dimension of linear layers & 2048 & Number encoder layers & 2 \\
Loss function & MSE & Early stopping patience & 3 \\
Decoder input size & 7 & Start token length & 48 \\
Use distilling & True & Activation function & gelu \\
Encoder input size & 7 & Batch size & 32 \\
Attention factor & 1 & Moving average window & 25 \\
Dimension of model & 512 & Maximum training epochs & 20 \\
Dropout & 0.05 & Output attention & False \\
Number of heads & 8 & Number decoder layers & 1 \\ \midrule
\multicolumn{2}{c}{Default LR schedule} & \multicolumn{2}{c}{Tuned LR schedule} \\ \midrule
LR schedule & type1 & LR schedule & cosine \\
Learning rate & 0.0004 & learning\_rate & 0.01 \\
 &  & lr\_k\_decay & 0.85 \\
 &  & min\_lr & 0.0004 \\
 &  & warmup\_lr & 0.0001 \\
 &  & warmup\_epochs & 3 \\ \midrule
\multicolumn{4}{c}{OIALR Hyperparameters} \\ \midrule
Delay & 600 & Full rank first layer & True \\
Full rank last layer & True & Stability frequency & 400 \\
Full rank warmup & False & Sigma cutoff fraction & 0.4 \\ \bottomrule
\end{tabular}
\end{table*}

\begin{table*}[h]
\caption{The search space and settings for the hyperparameter search using \texttt{Propulate}.}
\label{tab:af-prop}
\centering
\begin{tabular}{@{}llll@{}}
\toprule
Parameter & Search Space & Propulate Parameter & Value \\ \midrule
LR & (1e-5, 5e-3) & Crossover probability & 0.7 \\
Minimum LR & (5e-6, 1e-3) & Mutation probability & 0.4 \\
Warmup LR & (5e-6, 2e-4) & Random init probability & 0.1 \\
LR k-decay & (1e-3, 2) & Number of islands & 4 \\
Warmup epochs & (2, 10) & Migration probabilty & 0.9 \\
OIALR sigma cutoff fraction & (0.01, 0.9) &  &  \\
OIALR stability frequency & (50, 2000) &  &  \\
OIALR delay & (250, 2500) &  &  \\
OIALR full rank first layer & (False, True) &  &  \\
OIALR full rank last layer & (False, True) &  &  \\ \bottomrule
\end{tabular}
\end{table*}


\end{document}